\pdfoutput=1
\documentclass[11pt]{article}

\usepackage{acl}

\usepackage{times}
\usepackage{latexsym}
\usepackage{amsmath}
\usepackage{amssymb}
\usepackage[T1]{fontenc}

\usepackage[utf8]{inputenc}
\usepackage{microtype}
\usepackage{inconsolata}

\usepackage{xcolor}
\usepackage{soul}

\usepackage{enumitem}
\usepackage{graphicx}
\usepackage{booktabs}
\usepackage{multirow}
\usepackage{makecell}
\usepackage{amsthm}
\usepackage{stfloats}
\usepackage{forest}
\usepackage{fontawesome}
\usepackage{pifont}
\usepackage{amssymb}
\usepackage{array}
\usepackage{arydshln}
\usepackage{cleveref}

\usepackage{stfloats}

\title{A Survey on Natural Language Counterfactual Generation}

\author{Yongjie Wang\textsuperscript{1,*}, Xiaoqi Qiu\textsuperscript{2,*}, Yue Yu\textsuperscript{1}, Xu Guo\textsuperscript{1}, Zhiwei Zeng\textsuperscript{1},\\{\bf Yuhong Feng\textsuperscript{2,\dag}, Zhiqi Shen \textsuperscript{1,\dag} }   \\
   \textsuperscript{1} Nanyang Technological University, Singapore \\
   \textsuperscript{2} Shenzhen University, China\\ 
   \texttt{\textsuperscript{1}\{yongjie.wang,yue.yu,xu.guo,zhiwei.zeng,ZQshen\}@ntu.edu.sg} \\
   \texttt{\textsuperscript{2} qiuxiaoqi2022@email.szu.edu.cn, yuhongf@szu.edu.cn} 
   }

\begin{document}
\maketitle
\begingroup
\renewcommand\thefootnote{\fnsymbol{footnote}}
\footnotetext[1]{Equal contribution.}
\footnotetext[2]{Corresponding author.}
\endgroup
\begin{abstract}
Natural language counterfactual generation aims to minimally modify a given text such that the modified text will be classified into a different class. The generated counterfactuals provide insight into the reasoning behind a model's predictions by highlighting which words significantly influence the outcomes. Additionally, they can be used to detect model fairness issues and augment the training data to enhance the model's robustness. A substantial amount of research has been conducted to generate counterfactuals for various NLP tasks, employing different models and methodologies. With the rapid growth of studies in this field, a systematic review is crucial to guide future researchers and developers. To bridge this gap, this survey provides a comprehensive overview of textual counterfactual generation methods, particularly those based on Large Language Models. We propose a new taxonomy that systematically categorizes the generation methods into four groups and summarizes the metrics for evaluating the generation quality. Finally, we discuss ongoing research challenges and outline promising directions for future work.
\end{abstract}

\section{Introduction}

The recent advancements in Natural Language Processing (NLP) are driven by a variety of Large Language Models (LLMs), such as GPT-3 (175B) \cite{brown2020language}, PaLM (540B) \cite{chowdhery2023palm}, and GPT-4 (1.7T) \cite{achiam2023gpt}. These LLMs have demonstrated superior performance on various downstream tasks. However, alongside the performance, there is a rising concern about their occasionally undesired behaviors, like hallucinations in their responses \cite{ji-etal-2023-towards}, and misalignment with human expectations \citep{vafa2024large}. These phenomena coincide with the long-standing issue of training deep learning models, which were known to be vulnerable to spurious correlations with artifacts, shortcuts, and biases prevalent in real-world training data \cite{geirhos2020shortcut, hermann2020shapes}. Hence, there is a growing demand for LLM explainability to understand model decisions and enhance their robustness, particularly in high-stakes applications. 

Counterfactual generation has emerged as an effective way to probe and understand the reasoning behind the prediction of a model by highlighting which parts of the input influence the outcomes \citep{wachter2017counterfactual,miller2019explanation}. 
It makes minimal modifications to an original instance to create counterfactual examples (CFEs) with different predicted classes. CFEs can be used to detect model fairness issues within minority groups \citep{NIPS2017_a486cd07,NIPS2017_1271a702}, and enhance the robustness and generalizability of the model by augmenting the training dataset \citep{sen-etal-2021-counterfactually,wang2021robustness,qiu2024paircfr}.

In the field of NLP, early studies ~\cite{jung2022counterfactual,robeer2021generating} were inspired by traditional CFE generators for tabular data. 
However, due to the vast and discrete perturbation space of each word, directly applying these techniques in the NLP domain becomes less effective and inefficient. Additionally, textual CFEs should 
adhere to lexicon and grammar rules, and follow the language context and logic \cite{sudhakar2019transforming,wu-etal-2021-polyjuice,ross-etal-2021-explaining}. Subsequent research has begun to utilize the controlled text generation model to either rewrite a given sentence for the target label \cite{robeer2021generating,madaan2021generate} or replace influential words for the current prediction with alternatives for the target prediction \cite{ross-etal-2022-tailor,zhu-etal-2023-explain}. Until recently, the rise of LLMs has driven researchers to craft sophisticated prompts to obtain CFEs on a one-off basis \cite{chen2023disco,sachdeva-etal-2024-catfood}. 

As research on textual CFE generation expands rapidly, there is an urgent need for a systematic review specifically dedicated to this domain. However,
existing surveys in counterfactual generation primarily focus on tabular data \cite{verma2020counterfactual,stepin2021survey,10.1145/3527848,guidotti2022counterfactual,wang2023counterfactual}, and fail to offer comprehensive guidelines for researchers and developers within the NLP community.

The challenge of reviewing this area arises from the following factors. 
Firstly, the generation methods are inherently tied to the task definitions; different applications such as sentiment analysis and question answering require tailored generation strategies. 
Secondly, the formulation of the generation problem varies depending on the modification strategies and language models chosen. Finally, to fully comprehend and evaluate various algorithms, a comprehensive and interdisciplinary understanding that extends beyond NLP to include generative modeling, causality, and AI explanation is essential. This multidisciplinary requirement significantly enhances the complexity and challenge of conducting an exhaustive systematic review.

In this survey, we review past research on natural language counterfactual generation and categorize these methods into four groups: (1) Manual generation, where a human annotator is asked to edit a limited number of words for a given text to change its label \citep{kaushik2019learning}; (2) Joint learning-based generation involves training an end-to-end model that jointly minimizes the desired objectives using gradient descent \citep{robeer2021generating,yan2024counterfactual}; (3) Identify and then generate, a two-stage approach that pinpoints and then substitutes words to alter the labels \citep{malmi-etal-2020-unsupervised,gilo2024general,martens2014explaining}; and (4) LLMs prompting, which directly create the counterfactuals via prompting LLMs \cite{bhattacharjee2024llmguided,gat2023faithful,sachdeva-etal-2024-catfood}. 
We also summarize the qualitative and quantitative metrics used to evaluate the quality of the generated counterfactuals. Finally, we discuss the remaining challenges in this field and outline promising research directions, particularly in the era of LLMs. 

The rest of this paper is organized as follows: Section \ref{sec:definition of cfe} introduces the definition of CFEs and practical considerations during generation. Section \ref{sec:methods category} presents our novel taxonomy and describes each group. Section \ref{sec:evaluation metric} summarizes the metrics used to evaluate generation quality. Section \ref{sec:challenges} discusses ongoing challenges and promising research directions. Finally, Section \ref{sec:conclusion} concludes the paper.

\section{Definition of Counterfactual Example}
\label{sec:definition of cfe}
In machine learning, a counterfactual example (CFE), was initially proposed to explain model decisions on tabular data \cite{wachter2017counterfactual,miller2019explanation,verma2020counterfactual}. CFE explains why the model predicts an instance $\boldsymbol{x}$ as the class $y$ instead of its alternative $y^\prime$ by making \textit{minimal yet necessary} changes to $\boldsymbol{x}$ to obtain the desired change in its prediction. 

\begin{figure*}
    \centering
    \includegraphics[scale=0.6]{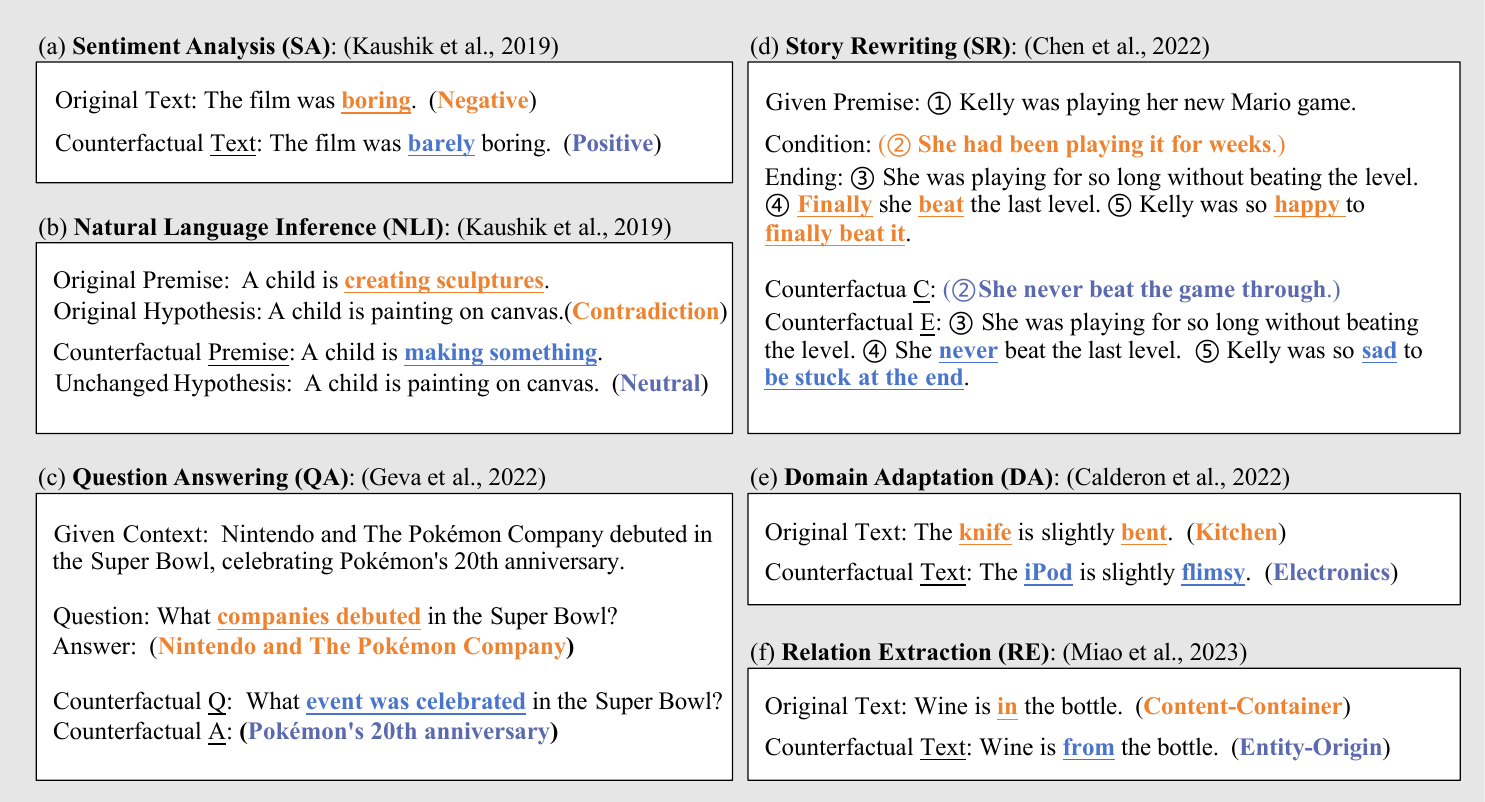}
    \caption{Use cases of counterfactual generation in various NLP tasks.}
    \label{fig:cf-egs-tasks}
\end{figure*}

We assume a trained model $f:\mathcal{X} \subset \mathbb{R}^d \to \mathcal{Y}$ is employed to predict the label of an input instance $\boldsymbol{x}$: $f(\boldsymbol{x}) = y$. $\mathcal{Y}$ represents a set of discrete labels for a classification task; whereas for a regression task, $\mathcal{Y}$ denotes a continuous real space. Given an input sentence $\boldsymbol{x}\in\mathcal{X}$ with its prediction $y$, a counterfactual generation method $g: f \times \mathcal{X} \to \mathcal{X}$ modifies a minimal subset of the words of $\boldsymbol{x}$ to produce a CFE $\boldsymbol{c}$, which alters the model's prediction to a desired class $y^\prime$: $f(\boldsymbol{c}) = y^\prime$, where $y^\prime\neq y$. Hence, generating counterfactual examples can be achieved by solving the following constrained optimization problem, 
\begin{align}
\label{eq:definition}
   & \operatorname*{\arg\min_{\boldsymbol{c}}} dist(\boldsymbol{x}, \boldsymbol{c}) \\
                            & \  s.t. \ f(\boldsymbol{c}) = y^\prime , \nonumber
\end{align}
where $dist(\cdot,\cdot)$ is a distance function that measures the cost of changes required to alter the prediction. To concisely define our research scope, we distinguish between two similar yet distinct terms: `adversarial examples' and `style transfer', in Section \ref{apd:terminology}, Appendix. 

This definition outlines the fundamental principles of the CFE generation problem, which can be adapted to a range of NLP tasks by specifying the task-specific $\boldsymbol{x}$ and $y^\prime$. For example, in question answering, the goal is to minimally revise the question $\boldsymbol{x}$ to satisfy a different answer $y^\prime$. In Figure \ref{fig:cf-egs-tasks}, we present examples of counterfactual generation across various NLP tasks. For detailed definition of CFE generation for those NLP tasks, please refer to Section \ref{apd:Cases of Natural CFE} in the Appendix. 

In practice, researchers typically impose various constraints to guide the generation of CFEs for specific objectives. Below, we outline the commonly accepted desiderata.

\noindent \textbf{Validity:} A CFE is valid if it correctly classified as the desired prediction. Optimizing validity will encourages a higher rate of successful label flips. 

\noindent  \textbf{Proximity:} It is the key constraint to create ``close possible worlds'' that preserve most of the original content while altering only the critical words to have a different prediction \cite{wachter2017counterfactual}. 

\noindent \textbf{Diversity:} A diverse set of CFEs contains multiple possible revisions of a sentence to achieve the target prediction where each revision reveals a different prediction logic. Such broad reasoning analysis enhances users' trust in a model's prediction \cite{wachter2017counterfactual}. A diverse set of CFEs also allows us to augment a model training for stronger robustness \cite{joshi-he-2022-investigation,qiu2024paircfr}. 


\noindent \textbf{Fluency:}  It measures the smoothness and naturalness of a CFE, similar to plausibility in tabular CFE generation \citep{gilo2024general}. Encouraging fluency results in texts that are grammatically correct, semantically meaningful, and coherent, which is crucial for ensuring that a textual CFE is understandable.

Recent research also include desiderata such as controllability \citep{ribeiro-etal-2020-beyond, wu-etal-2021-polyjuice} and stability \cite{gardner-etal-2020-evaluating, geva2022break} to better control or stabilize the generation process. However, these desiderata do not directly describe the desired format of the final CFEs \cite{guidotti2022counterfactual}. Due to page limit, detailed discussion is omitted. 

\section{Counterfactual Generation Methods}
\label{sec:methods category}

In this section, we carefully collect $66$ studies for textual counterfactual generation. The detailed process of collection is described in Section \ref{apd:paper_collection} of Appendix. After that, we propose a novel taxonomy that categorizes existing methods into four groups. Within each group, we further divide these methods into fine-grained subgroups or successive steps, to ensure that the taxonomy is systematically organized. The full taxonomic structure is shown in Figure \ref{fig:full_tree} of the Appendix.

\subsection{Manual Generation}
\label{sec:manual gcf}
Generating high-quality textual CFEs has proven to be challenging for neural networks. Consequently, early studies relied on domain experts or crowdsourcers to manually collect these CFEs \cite{kaushik2019learning,gardner-etal-2020-evaluating,yang-etal-2020-generating,samory2021call}.

Before editing, human annotators are given detailed instructions and examples. The editing principles include: (1) Minimal Edits: 
using domain knowledge to minimally edit the original text, such as deletion, insertion, replacement, and reordering. (2) Fluency, Creativity, and Diversity: ensure that edits maintain fluency and grammatical accuracy, while also introducing diverse modifications, including changes to adjectives, entities, and events. (3) Adhere to task-specific rules. For instance, in question-answering (QA) tasks, counterfactual questions should be answerable based on the given context \cite{khashabi-etal-2020-bang}.

To improve revision quality, multiple annotators are often employed to cross-validate the revised CFEs \cite{kaushik2019learning, gardner-etal-2020-evaluating}. Those with lower consensus are then filtered out. However, creating a high-quality CFE dataset through human labor is both time-consuming and expensive \cite{sen-etal-2023-people}. For instance, \citet{kaushik2019learning} reported that modifying and verifying a single CFE typically takes four to five minutes and costs approximately \$0.8. 

\subsection{Joint Learning-based Generation}
\label{sec:gradient}

The constrained problem in Equation \eqref{eq:definition} can be converted to the Lagrange function below,
\begin{align}
    \label{eqn:lagrange}
    \mathcal{L} = dist(\boldsymbol{x}, \boldsymbol{c}) + \lambda_1 \cdot \ell(f(\boldsymbol{c}), y^\prime),
\end{align}
where $\ell(\cdot, \cdot)$ describes the difference between the desired target $y^\prime$ and current prediction $f(\boldsymbol{c})$, and $\lambda_1 \in \mathbb{R}^+$ is the Lagrange multiplier. A larger $\lambda_1$ will encourage the CFEs to be closer to the desired prediction. 
Additional desired properties or constraints, such as diversity and fluency, can also be formulated using corresponding mathematical functions, which are appended after Equation \eqref{eqn:lagrange}. 

Neural networks such as, BERT \cite{devlin2019bert}, GPT-2 \cite{radford2019language}, are differentiable, and the distance function, denoted as $dist(\cdot)$, typically employs either the $L_1$ or $L_2$ norm.
Consequently, researchers \cite{madaan2021generate,hu2021causal,jung2022counterfactual} can employ gradient descent to iteratively minimize the joint loss until specific stopping conditions are met.

Optimizing the joint loss in Eqn. \eqref{eqn:lagrange} for a specific sentence $\boldsymbol{x}$ could help find its CFEs \cite{jung2022counterfactual}, but this approach invokes the model multiple times for each input, not efficient.  Therefore, a family of research \cite{madaan2021generate, hu2021causal, madaan2023counterfactual, yan2024counterfactual} directly learned a counterfactual generation model by optimizing the joint loss over a collection of annotated CFEs. During inference, the generation model is fed the input text and its target class, and it directly returns a CFE that belongs to the target class. These generation models are built under the following two frameworks: 

\noindent(1) Controlled text generation framework (PPLM) \cite{Dathathri2020Plug}. It combines a frozen language model with additional small attribute models that guide the generation towards specific themes, emotions, or styles of writing. In particular, the attribution model is trained to perturbs the hidden states of input texts to maximize the desired characteristics. The follow-up studies  GYC \cite{madaan2021generate} and 
CASPer \cite{madaan2023counterfactual} utilize the PPLM framework for counterfactual generation. 

\noindent(2) VAE and GAN frameworks. CounterfactualGAN \cite{robeer2021generating} uses the StarGAN \cite{8579014} to ensure that the CFEs adhere to the data distribution. To prevent model learning spurious correlations, \citet{hu2021causal} develop a causal model using variational auto-encoders (VAEs). \citet{yan2024counterfactual} disentangle content and style representations using a VAE model. They then intervene in the style variable while maintaining the content variable constant, enabling the generation of counterfactual explanations through the decoder model.

As these generation models are trained in an end-to-end manner, one limitation is that we cannot concisely control the generation process, such as which words should be revised. Additionally, the model is trained by minimizing the loss over a collection of samples, which may compromise the quality of CFEs for certain sentences, such as those pertaining to minority groups. Lastly, controlled text generation does not necessarily produce CFEs with minimal and diverse perturbations. 







\subsection{Identify and then Generate}
\label{sec:2-step}

\begin{figure*}
    \centering
    \includegraphics[width=0.9\textwidth, keepaspectratio]{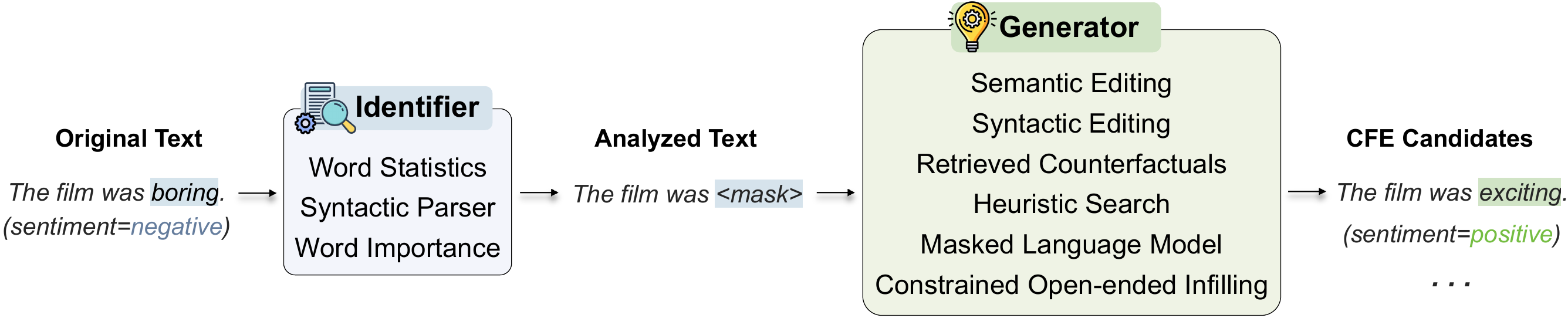}
    \caption{Demonstration of the Identify-and-then-Generate CFE generation.}
    \label{fig:2step}
\end{figure*}

A popular family of approaches decomposes the generation task into two steps: (1) identifying the words to be revised in the original text, and (2) minimally editing those words to generate CFE candidates with target predictions, as shown in Figure \ref{fig:2step}. 

\subsubsection{Identification step}
\label{subsec:identify}

The simplest strategy involves either selecting random words \cite{fu-etal-2023-scene} or revising all words \cite{fern-pope-2021-text}. However, such approaches fail to discriminate between words that potentially contribute to valid counterfactuals and those that do not. Consequently, the subsequent generation step may produce futile results, leading to unnecessary costs. Therefore, researchers propose more deliberately designed identifiers, which are summarized as follows: 

\noindent \textbf{(1) Words statistics.} This approach \cite{madaan-etal-2020-politeness,li-etal-2018-delete} first calculates the frequency of words or n-grams that appear in the target domain corpus using traditional term frequency (TF) and/or inverse document frequency (IDF) measures. It then marks those words or n-grams whose frequency scores exceed a specific threshold.

\noindent \textbf{(2) Syntactic parser.} Syntax plays a crucial role in model predictions across many tasks. For example, adjectives (`good', `delicious') and verbs (`like', `hate') are often considered closely linked to sentiment polarity. Subjects and objects are important for understanding the logical relationship in the NLI task. Consequently, researchers \cite{chen2023disco,geva2022break} adopt a syntactic parser to split a sentence into spans. Control codes \cite{ribeiro-etal-2020-beyond,wu-etal-2021-polyjuice} are incorporated into parsers to produce different types of perturbations for various purposes. Additionally, Tailor \cite{ross-etal-2022-tailor} analyzes text syntax to extract high-level and semantic control codes, enabling flexible and meaningful perturbation strategies.

\noindent \textbf{(3) Word importance.} The approaches in this category identify important words that significantly contribute to the original prediction. For example, given a positive text such as ``It is a fantastic moment,'' the word `fantastic' would be identified as the crucial word for the positive label. Compared to identifiers based on word statistics and syntactic parsers that only require an input sentence, the word importance-based identifier additionally necessitates a pretrained model to judge word importance via prediction differences.

Conveniently, importance scores can be readily obtained from current feature importance approaches such as gradients \cite{DBLP:journals/corr/SimonyanVZ13}, integrated gradients \cite{pmlr-v70-sundararajan17a}, LIME \cite{ribeiro2016should}, SHAP \cite{lundberg2017unified}, and CURE \cite{si2023consistent}. For instance, MICE \cite{ross-etal-2021-explaining} uses gradients to determine which tokens to mask; LEWIS \cite{reid-zhong-2021-lewis} identifies style-related tokens with above-average attention weights; Polyjuice \cite{wu-etal-2021-polyjuice} and AutoCAD \cite{wen2022autocad} incorporate LIME and SHAP as plugins to identify mask positions; \citet{martens2014explaining} identify a minimal set of words whose removal would revert the current prediction. Typically, a higher importance score indicates a greater significance to the original prediction, and such tokens are more likely to be replaced in the generation step.  \qed



The above techniques can be combined to achieve more precise identification of editing locations. For instance, AC-MLM \cite{ijcai2019-732} combines word frequency and attention scores to obtain accurate locations. 

For word statistics and word importance-based identifiers, each word is assigned a score. Then, we need to determine how many words should be masked. Masking too many words compromises CFE’s proximity, while masking too few may result in void CFEs.  Empirically, recent studies often employ predefined rules, such as selecting the top-K words or spans \cite{malmi-etal-2020-unsupervised,wen2022autocad},  choosing words whose scores exceed a certain threshold \cite{ijcai2019-732,hong-etal-2023-robust}, or adaptively controlling the number of masked tokens \cite{reid-zhong-2021-lewis,madaan-etal-2020-politeness}. 

 
\subsubsection{Generation step}
\label{subsec:generate}
Once the words to be revised are identified, the next step is to replace the these words with appropriate alternatives to achieve the target prediction. We list common generation methods below:

\noindent \textbf{(1) Semantic editing}. An intuitive solution is to substitute the important words with their corresponding semantic counterparts such as antonyms. They can be readily obtained with existing lexical databases like WordNet \cite{chen2021kace,wang2021robustness,chen-etal-2021-reinforced}. Alternatively, they can be searched within the dataset of the target class \cite{li-etal-2018-delete,gilo2024general}. This strategy is limited to tasks related to semantic understanding \cite{wen2022autocad}.


\noindent \textbf{(2) Syntactic editing.} These methods \cite{li-etal-2020-linguistically,zhu-etal-2023-explain,longpre-etal-2021-entity,geva2022break} leverage existing language parsers to decompose a sentence into several syntactic spans, then design customized rules to transform each span into the desired output. Examples include inserting `not' before verbs or adjectives, swapping subjects and objects, modifying tense, substituting a word with another entry from the corpus, or tampering with factual evidence. Such approaches are primarily designed for tasks like natural language inference, named entity recognition, and fact verification, where the model predictions are sensitive to the tense, location of passive and subject, and evidence.



\noindent \textbf{(3) Retrieved counterfactuals.}  Retrieval-based approaches \cite{li-etal-2018-delete} first retrieve an open-source database using the masked original sentence. Subsequently, filtering techniques are used to keep valid and minimally revised candidates. RGF \cite{paranjape-etal-2022-retrieval} directly generates counterfactual questions based on retrieved context and answers in QA task. Although RGF does not need to identify word positions, we categorize this method here due to its use of retrieval techniques.  The major concern with this approach is that the retrieved counterfactuals may not be as similar to the original sentence as other methods. 

\noindent \textbf{(4) Heuristic search.} These methods \cite{fern-pope-2021-text, gilo2024general} employ heuristic search to find appropriate replacements within a defined search space. The key contributions of these methods are the construction of the search space and the development of search strategies. \citet{fern-pope-2021-text} first identify the $k$ potential substitutions for each word and adopt a Shapley-value guided search method. \citet{gilo2024general} start from a CFE in the training dataset and leverage the weighted $A^*$ algorithm to iteratively reduce the edit cost.

\noindent \textbf{(5) Masked language models (MLMs).} The identified word locations can be masked with specific tags such as `[MASK]'. An MLM can then be used to edit these tags to achieve the target prediction. For example, consider a masked sentence like ``There is a [MASK] moment,'' with a goal to generate a negative expression, MLMs might fill in the mask with words like `terrible' or `dismal'. 

The primary contributions of approaches in this family revolve around how they leverage and train MLMs for infilling tasks. (1) Some methods \cite{ribeiro-etal-2020-beyond,chen2022unsupervised,chemmengath-etal-2022-cat} directly leverage the pretrained MLMs to infill the blanked words. While convenient, the generated words may not always align with the desired properties, often necessitating post-hoc filtering to meet user expectations.  (2) Other approaches finetune MLMs on target domain data \cite{malmi-etal-2020-unsupervised,reid-zhong-2021-lewis} and then use finetuned MLMs to infill the blanks. (3) A widely adopted method \cite{ijcai2019-732,ross-etal-2021-explaining,hao2021sketch,calderon-etal-2022-docogen,wen2022autocad} involves finetuning the MLM to realize label-controlled generation from the masked sentences and their conditional label. Here, the MLM learns to infill the blank that is consistent with the conditional label. (4) Some researchers directly finetune an MLM to learn the counterfactual generation from the masked sentence to desired formats \cite{wu-etal-2021-polyjuice,ross-etal-2022-tailor}. However, this approach often requires a substantial amount of training data. For instance, \cite{wu-etal-2021-polyjuice} recommends collecting 10,000 instances per control code, which can be burdensome. 

The primary drawback of these approaches is that MLMs focus solely on revising the masked positions, which leads to a lack of linguistic diversity in generated CFEs.

\noindent \textbf{(6) Constrained open-ended infilling.} This approach aims to infill the masked positions more flexibly while restricted by a label flip rate constraint, compared to MLM approaches that strictly infill the mask locations with replacements. For example, NeuroCFs \cite{howard-etal-2022-neurocounterfactuals} first identify key concepts and then use a GPT-2 model, adapted to the target prediction, to decode these concepts. DeleteAndRetrieve \cite{li-etal-2018-delete} concatenates the embeddings of the masked original sentence and a retrieved sentence with the target prediction, then adopts a decoder to generate a CFE.

\subsection{LLMs Prompting}
 
 \label{sec:prompting}

In the past two years, LLMs have shown remarkable proficiency in synthesizing natural languages for downstream tasks \citep{meng2022generating,ye-etal-2022-zerogen,meng2023tuning,yu2024large}.
Significant research has focused on designing effective prompts to harness the advanced reasoning and understanding capabilities of these models for generating desired content, including CFEs \citep{dixit-etal-2022-core,gat2023faithful,chen2023disco}. In recent literature, two key technologies in enhancing the generation results are In-Context Learning (ICL) and Chain-of-Thought (CoT).

Introduced with GPT-3 \cite{brown2020language}, ICL improves prompts by including examples that demonstrate the expected type of reasoning or output. To generate counterfactuals for a given instance, the prompt typically consists of the task requirement and one \citep{sachdeva-etal-2024-catfood} or a few pairs of original and counterfactual examples as demonstrations \cite{dixit-etal-2022-core,chen2023disco,gat2023faithful,sachdeva-etal-2024-catfood}. These in-context counterfactuals are either manually created \cite{chen2023disco,gat2023faithful} or retrieved from an external unlabeled corpus \cite{dixit-etal-2022-core}.

CoT prompting, introduced by \citet{wei2022chain}, elicits the emergent reasoning capability of LLMs by incorporating a series of intermediate reasoning steps into the prompt. 
For example, in sentiment classification, generating counterfactuals for a positive sentence involves two steps: (1) identifying and (2) altering words that convey positive sentiment \cite{bhattacharjee2024llmguided,nguyen2024llms,li2024prompting}.
This technique is more evident in question-answering tasks, where \citet{sachdeva-etal-2024-catfood} demonstrate that the counterfactuals for an answer can be obtained by first generating a counterfactual question based on the factual context and then producing the corresponding answer. 

\begin{table*}[htbp]
\centering
\caption{{Summary of the four categories of natural language CFEs generation.}}
\label{tab:four_cat_comparison}
\resizebox{\textwidth}{!}{%
\begin{tabular}{m{1.8cm}|p{3.8cm}|p{5.5cm}|p{5.0cm}|p{3.5cm}}
\hline
\textbf{Category} & \textbf{Human Generation} & \textbf{Joint Learning-based Generation} & \textbf{Identify and then Generate} & \textbf{LLMs Prompting} \\ \hline
\textbf{Description} & Instructing human annotators to revise a sentence & Training an end-to-end model that jointly minimizes the multiple objectives associated with user desiderata. & Employing a divide-and-conquer strategy, identifying important words and replacing them with alternatives & Prompting LLMs to generate CFEs \\ \hline
\textbf{Training} & No & Yes & Optional & No \\ \hline
\textbf{Pros} & Meaningful and minimal revision, high quality & End-to-end, quantifiable objectives; easy to optimize the joint objective & Explainability; high controllability; precise edit & User-friendly; cheaper than human; no training \\ \hline
\textbf{Cons} & Time-consuming; labor-intensive; expensive & Hard to quantify each objective; trade-off over multiple objectives; lower controllability  & Complicated workflow & Hard to tune prompts; rely on prompt quality \\ \hline
\end{tabular}%
}
\end{table*}

\subsection{Filter}
\label{sec:filter}

Since the automatic counterfactual generators may produce degenerate counterfactuals (incoherent, illogical, or invalid) for some input texts, post-hoc filtering is typically employed to filter out these degenerate cases.

Human filtering \cite{zhang-etal-2019-paws} ensures high-quality CFEs but it is time-consuming and labor-intensive. Therefore, researchers use automated tools to remove undesired outputs. These automated methods include eliminating CFE candidates that are incorrectly predicted by state-of-the-art (SOTA) models \cite{reid-zhong-2021-lewis,zhang2023towards,chang2024counterfactual}; deleting degenerations with low fluency scores computed by language models \cite{li-etal-2020-linguistically,wu-etal-2021-polyjuice,ross-etal-2022-tailor,gilo2024general}; and selecting human-like counterfactuals based on proximity scores \cite{yang-etal-2021-exploring}. 

\subsection{Summary}

We summarize the characteristics, strengths, and weaknesses of each category of natural language CFE generation approaches in Table \ref{tab:four_cat_comparison}.
Owing to page limit, we only discuss a few of the most pertinent studies for each category to ensure that the essential information is conveyed clearly. Complete discussion of relevant references for each category can be found in Appendix Section \ref{apd:paper list}.

\section{Evaluation Metrics}
\label{sec:evaluation metric}
\textbf{Validity.} It measures the proportion of CFEs that achieve the desired target among all generated CFEs. Formally, the validity over $N$ test samples is defined by,
\begin{align}
    Validity = \frac{1}{N} \sum_{i=1}^{N}\mathbb{I}(\hat{f}(\boldsymbol{c}_i) = y_i'),
\end{align}
where $y_i'$ is the desired target of a CFE $\boldsymbol{c}_i$. The predictor $\hat{f}$ can be human annotation \cite{wu-etal-2021-polyjuice,chen2021kace}, fined-tuned SOTA models (e.g., RoBERTa~\cite{ross-etal-2021-explaining,wen2022autocad,betti2023relevance,balashankar2023improving,gat2023faithful}, or BERT~\cite{betti2023relevance,bhattacharjee2024llmguided} in sentiment analysis, and DeBERTa~\cite{chen2023disco} in natural language inference), or voting with multiple models \cite{sachdeva-etal-2024-catfood}. A higher validity is preferred. 

\textbf{Similarity.} Similarity measures the editing effort required of a CFE during generation \cite{wu-etal-2021-polyjuice,kaushik2019learning}, formally defined as, 
\begin{align}
    Similarity = \frac{1}{N} \sum_{i=1}^{N} dist(\boldsymbol{x}_i, \boldsymbol{c}_i).
\end{align}
For lexical and syntactic similarity evaluations, widely used methods include the word-level Levenshtein edit distance~\cite{levenshtein1966binary} and the syntactic tree edit distance~\cite{zhang1989simple}. For assessing semantic similarity, models \footnote{pretrained models and not finetuned on evaluation tasks.} like SBERT~\cite{reimers2019sentence} and the Universal Sentence Encoder (USE)~\cite{cer-etal-2018-universal} are commonly used. They encode both the CFE and the input text and then calculate the cosine similarity between their sentence representations.

\textbf{Diversity.}  This score is measured as the average pairwise distance between $K$ returned CFEs for a sentence $\boldsymbol{x}$, defined as follows, 
\begin{align}
    Diversity = \frac{1}{\binom{K}{2} }\sum_{i=1}^{K-1} \sum_{j=i+1}^K dist(\boldsymbol{c}_i,\boldsymbol{c}_j).
\end{align}
For lexical diversity, Self-BLEU \cite{zhu2018texygen} reports the average BLEU score between any two CFEs, while Distinct-n \cite{li-etal-2016-diversity} gauges diversity by calculating the ratio of unique n-grams to the total number of n-grams in the generated CFEs. When semantic diversity is assessed, the $dist(\cdot)$ function can be metrics like SBERT embedding similarity~\cite{reimers2019sentence}, BERTScore~\cite{zhang2019bertscore}, semantic uncertainty~\cite{kuhn2023semantic}.

\textbf{Fluency.} As fluency describes the resemblance of a CFE to human writing, a simple measurement is to ask human raters to evaluate a CFE based on cohesiveness, readability, and grammatical correctness \cite{robeer2021generating,madaan2021generate}. Due to the irreproducibility and high cost of human evaluation, automated fluency evaluations such as the likelihood and the perplexity score have become popular in recent studies \cite{ross-etal-2021-explaining,sha-etal-2021-controlling,treviso-etal-2023-crest}.

(1) \textit{Likelihood}~\cite{salazar-etal-2020-masked}. Given a sentence of length $n$, we create $n$ copies by individually masking each of the $n$ tokens. We then use a masked language model (MLM), such as T5-based models, to compute the loss for both the original sentence and its $n$ masked copies. The likelihood is calculated as the average ratio of the loss of each masked copy to the loss of the original sentence.
  
(2) \textit{Perplexity score} \cite{10.1121/1.2016299}. This score evaluates whether the produced CFEs are natural, realistic, and plausible. In practice, we quantified this using the powerful generative LMs (e.g., GPT-2 \cite{radford2019language}), formally described as follows,
\begin{align}
    perplexity =\mathrm{exp} \! \left [ -\frac{1}{n}  \sum_{i=0}^{n} \log p_{\theta }(t_i|t_{<i})\right ],
\end{align}
where $p_{\theta }(t_i|t_{<i})$ is the probability of the $i$-th token of a CFE $\boldsymbol{c}$, given the sequence of tokens ahead.

\textbf{Model Performance.} 
As revision in CFEs ideally reveal important features, we can either incorporating CFEs into training to enhance model robustness \cite{chen2021kace,qiu2024paircfr} or leverage CFEs as test sets to evaluate existing model's generalization \cite{ribeiro-etal-2020-beyond, ross-etal-2021-explaining}. Researchers then report the classification performance, such as accuracy, F1-score, and the standard deviation of these metrics on out-of-domain datasets or counterfactual test sets. 

The evaluation mentioned above can also be conducted by human evaluators where humans are instructed to rate CFEs from various aspects \cite{ijcai2019-732,madaan2021generate}. In Appendix Section \ref{apd:summary_evaluation}, we summarize the commonly used evaluation metrics. Additionally, we also list the evaluation metrics used in experiments of each paper in Appendix Section \ref{apd:paper list}.

\section{Challenges and Future Directions}
\label{sec:challenges}

\textbf{Fair evaluation.} The absence of ground truth makes it difficult to compare CFEs generated from different methods. This challenge arises from two main aspects: (1) Existing metrics evaluate CFEs from various, often non-comparable perspectives. For example, prioritizing higher proximity (minimal changes to the original text) typically results in lower flip rate. Optimizing one metric may compromises another metric, making it difficult to dominate across all metrics and conclusively identify the best method. (2) Many methods use filtering techniques to discard undesired results. Direct comparisons between filtered and unfiltered CFEs may introduce bias in the evaluation process. For instance, methods employing GPT-2 to filter out grammatically incorrect or nonsensical sentences \cite{radford2019language,ross-etal-2022-tailor} often outperform those that do not use such filters on fluency score. 

\noindent \textbf{Model privacy and security.} Privacy and security are crucial considerations in the model development and deployment. As CFEs reveal sensitive changes near the decision boundary, researchers exploited CFEs to efficiently extract high-fidelity surrogate models \cite{aivodji2020model,10.1145/3531146.3533188}, which poses high risks to model privacy and security. Future research should focus on strategies to mitigate model extraction risks while maintaining the utility of CFEs.


In recent years, there has been an increasing trend toward using LLMs to generate counterfactuals. Following this, we outline and discuss the research challenges associated with prompting LLMs.

\noindent \textbf{Long-context CFEs generation.} Although LLMs can produce fluent and reasonable CFEs, our empirical studies reveal that when input sentences become longer, the quality of the generated CFEs quickly deteriorates. Even when running within the maximum token limit, LLMs can produce CFEs that are invalid, truncated, and overly summarized. Future work should investigate the generation of CFEs for long-context inputs.

\noindent \textbf{Hard to improve CFE quality.} With the aid of ICL and CoT prompting, LLMs can produce high-quality CFEs. However, it is still unclear which specific prompts are crucial for enhancing CFE quality. Although we observe certain issues, they do not offer clear guidance on how to address them. We should cultivate a deeper understanding of LLMs and strategically design prompts to target and resolve specific issues during CFE generation. 

\noindent \textbf{Specific LLMs for CFEs.} Modern LLMs are primarily trained on autoregressive tasks and then are fine-tuned with human feedback to enhance their ability to follow instructions. The commonly used tasks for instruction tuning are question answering and semantic understanding. The LLM potential of CFE generation may not be fully exploited during fine-tuning. We believe that fine-tuning LLMs specifically for the CFE generation task could enhance their performance.

\noindent \textbf{LLM hallucinations.} LLMs can generate incorrect, misleading, or entirely fabricated content with high confidence, a phenomenon formally known as LLM hallucination. When counterfactual data is used as ground truth to test or improve model robustness, this hallucinated content can inject misleading and incorrect relationships. Therefore, we should implement post-processing and fact-checking techniques to filter out hallucinated content by verifying against known facts and identifying internal contradictions. 

\noindent \textbf{Lower controllability.} LLMs may not always effectively determine the degree of change or the specific elements that should be altered in a given sentence, even with clear instructions. Without fine controllability, we cannot achieve the diversity that is possible when instructing human annotators. A nuanced understanding of LLM internal mechanisms is necessary to generate CFEs both flexibly and effectively. 

\section{Conclusion}
\label{sec:conclusion}

In this survey, we systematically review recent advancements, including the latest LLM-assisted generation approaches. Based on algorithmic differences, we propose a novel taxonomy that categorizes these methods into four groups, providing an in-depth comparison, discussion, and summary for each group. Additionally, we summarize the commonly used metrics to evaluate the quality of counterfactuals. Lastly, we discuss research challenges and aim to inspire future directions. 

With the widespread use of black-box LLMs, issues such as explanations, fairness, and robustness have gained increasing attention. We believe this survey can serve as a comprehensive guideline to inspire future advancements that address these concerns.


\section{Limitations}
\label{sec:limitations}
While this survey provides a systematic overview of counterfactual generation in the NLP domain, it has several limitations. Firstly, this survey predominantly focuses on generating CFEs, but it omits extensive descriptions like counterfactual thinking or reasoning from cognitive psychology and philosophy, which could help readers understand the necessity of CFE generation. Secondly, although counterfactual generation in NLP intersects with fields like causality, linguistics, and social sciences, this survey centres on NLP-specific aspects and may not fully explore these interdisciplinary connections, potentially limiting a deeper understanding in those areas. Lastly, although this survey acknowledges that counterfactual generation offers several benefits such as enhancing explainability, model debugging, and training data augmentation, it does not delve deeply into how CFEs function in these scenarios. Understanding these impacts is crucial for researchers deploying CFEs in real-world applications.

\section*{Acknowledgements}
This research is supported, in part, by the WeBank-NTU Joint Research Institute on Fintech, Nanyang Technological University, Singapore. This research is also supported, in part, by the National Research Foundation, Prime Minister’s Office, Singapore under its NRF Investigatorship Programme (NRFI Award No. NRF-NRFI05-2019-0002). Any opinions, findings and conclusions or recommendations expressed in this material are those of the author(s) and do not reflect the views of National Research Foundation, Singapore. Xu Guo thanks the Wallenberg-NTU Presidential Postdoctoral Fellowship. Zhiwei Zeng thanks the support from the Gopalakrishnan-NTU Presidential Postdoctoral Fellowship. We also appreciate the support from the Shenzhen Science and Technology Foundation (JCYJ20210324093212034, 20220810135520002); Guangdong Province Undergraduate University Quality Engineering Project (SZU Academic Affairs [2022] No. 7) and Key Laboratory (2017B030314073).



\bibliography{custom}

\appendix

\clearpage

\section{Terminology Clarification}
\label{apd:terminology} 
In this section, we clarify two terms related to Counterfactual Examples (CFEs) to ensure a precise review scope.

\noindent \textbf{Adversarial Example v.s. CFE}. Both text adversarial examples \cite{li2020bert, garg2020bae} and CFEs aim to change model predictions with minimal modifications. However, adversarial examples are designed to deceive human perception, altering only the model's prediction without necessarily being human-perceivable as different.
In contrast, CFEs should ideally change both human and model predictions simultaneously. 

\noindent \textbf{Style Transfer v.s. CFE}. Style transfer \cite{sudhakar-etal-2019-transforming, hu2017toward} aims to revise the input sentence to achieve a target style. Unlike CFE generation, which sought for minimal perturbations, style transfer may involve complete sentence modifications to ensure the sentence conforms to the target style. However, when minimal perturbation is also required in some style transfer research, we treat both tasks the same and include these studies. 

\section{CFE Generation in NLP Tasks}
\label{apd:Cases of Natural CFE}
\begin{figure}
    \centering
    \includegraphics[scale=0.6]{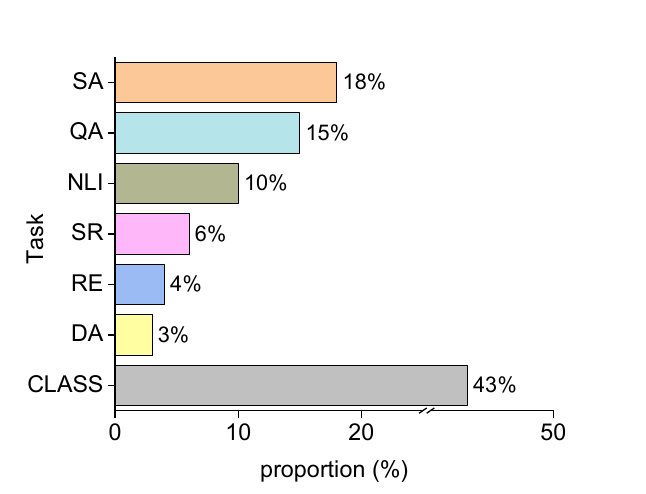}
    \caption{Proportion of papers in each task among all collected papers.  The term `CLASS' refers to papers applicable to general text classification tasks, including SA and NLI.}
    \label{fig:task proportion}
    \vspace{-0.4cm}
\end{figure}

Here, we present the formulation of CFE generation across various NLP tasks. In Figure \ref{fig:task proportion}, we report the proportion of papers in each task relative to all collected papers. 


\noindent \textbf{Sentiment Analysis (SA)} involves
determining the emotional polarity $y$ given a text $\boldsymbol{x}$. Counterfactual generation in SA refers to minimally modify the input text $\boldsymbol{x}$ such that the new sentence $\boldsymbol{c}$ has a different prediction $y'$, i.e., $(\boldsymbol{x},y) \to (\boldsymbol{c}, y')$.


\noindent \textbf{Natural Language Inference (NLI)} is to determine whether a given hypothesis $\boldsymbol{x}_1$ can be inferred from a given premise $\boldsymbol{x}_2$, and return a logical relationship $y$. CFE generation in NLI aim to revise hypothesis or premise or both to change current logical relationship $y$ to another different relationship $y'$, i.e., $(\boldsymbol{x}_1, \boldsymbol{x}_1, y) \to (\boldsymbol{c}_1, \boldsymbol{c}_2, y')$.


\noindent \textbf{Question Answering (QA)} aims to automatically produce an answer $\boldsymbol{a}$ for a given question $\boldsymbol{q}$ and context $\boldsymbol{x}$. The counterfactual QA task seeks to minimally modifies either the context  or the question, or both to generate counterfactual context $\boldsymbol{c}_x$ or question $\boldsymbol{c}_q$ such that $(\boldsymbol{c}_x,\boldsymbol{c}_q, \boldsymbol{a}')$ holds for a different answer $\boldsymbol{a}'$, i.e., $(\boldsymbol{x},\boldsymbol{q}, \boldsymbol{a}) \to (\boldsymbol{c}_x,\boldsymbol{c}_q, \boldsymbol{a}')$. 

\noindent \textbf{Story Rewriting (SR)}. The example in SR task includes a $5$ sentence tuple $\{\boldsymbol{s}_1, \boldsymbol{s}_2, \boldsymbol{s}_3, \boldsymbol{s}_4, \boldsymbol{s}_5 \}$ where $\boldsymbol{s}_1$ is the story premise, $\boldsymbol{s}_2$ is the initial context, and $\boldsymbol{s}_{3-5}$ are original story endings.  Given a contrastive context $\boldsymbol{s}_2'$, counterfactual SR aims to minimally revise the original endings, such that the revised endings $\boldsymbol{s}_{3-5}'$ still keep narrative coherency to the new context and original premise.


\noindent \textbf{Domain Adaptation (DA)}. Given a sentence $\boldsymbol{x}$ that belongs to the source domain $\boldsymbol{d}_s$, counterfactual DA aims to minimally intervene the original sentence such that the edited sentence $\boldsymbol{c}$ belongs to a different target domain $\boldsymbol{d}_t$.


\noindent \textbf{Relation Extraction (RE)} involves extracting the relationship $\boldsymbol{r}$ between entities in a given sentence $\boldsymbol{x}$. In counterfactual RE, we aim to minimally revise the $\boldsymbol{x}$ such that a different relationship $\boldsymbol{r}'$ can be obtained between these entities from the revised sentence $\boldsymbol{c}$.


\section{Paper Collection}
\label{apd:paper_collection}

This section outlines the approach we employed to collect relevant papers in this survey. We first retrieve papers from arXiv and Google Scholar with keywords ``counterfactually augmented data'', ``counterfactual explanation'', ``counterfactual generation'',``contrast set'', and ``contrastive explanation'', and finally we obtain over $200$ publications. We then filter out papers that merely apply CFE on specific applications or generally discuss CFE, retaining approximately 40 papers as our seed references. We then applied backward and forward snowballing techniques, examining the references and citations of these seed papers to identify additional relevant studies. We carefully reviewed all identified papers, focusing on those introducing novel counterfactual generation methods, which finally form this survey.


Our research paper list is available on GitHub \footnote{https://github.com/Siki-cloud/Awesome-CF-Generation}





\section{Summary of Evaluation Metrics}
\label{apd:summary_evaluation}
The evaluation metrics for comparing different CFEs are summarized in Table \ref{tab:full metric}. Here, we only list metrics that have been used in at least three publications.

\section{Summary of CFE Generation}
\label{apd:paper list}
In this section, we summarize all collected papers for each group in Section \ref{sec:methods category}. Due to the distinct characteristics of different method groups, we organized them into four separate tables, rather than merging them into one large table. For methods within a table, we can conveniently understand a method or compare it with another. The detailed summary are shown in Table \ref{tab:human paper list}, Table \ref{tab:gradient paper list}, Table \ref{tab:llm paper list}, and Table \ref{tab:ig paper list}.

\begin{table}[t]
\caption{Commonly used metrics for evaluating counterfactuals, where $\uparrow$ ($\downarrow$) indicates higher (lower) scores are better, and ($\rightarrow 1$) indicates closer to 1 is better.}
\label{tab:full metric}
\tabcolsep=0.1cm
\renewcommand\arraystretch{1.2}
\centering
\resizebox{\linewidth}{!}{%
\begin{tabular}{lclc}
\Xhline{1.2pt}
\multicolumn{2}{l}{Property} & Metric & Trend  \\ \hline
\multicolumn{2}{l}{Validity} & Flip Rate & $\uparrow$  \\ \hline
\multirow{8}{*}{Proximity} & \multirow{5}{*}{Lexical} & BLEU~\cite{papineni-etal-2002-bleu} & $\uparrow$   \\
 &  & ROUGE~\cite{lin-2004-rouge} &  $\uparrow$    \\
 &  & METEOR~\cite{denkowski-lavie-2011-meteor} &  $\uparrow$   \\
 &  & Levenshtein Dist.~\cite{levenshtein1966binary} & $\downarrow$    \\
 &  & Syntax Tree Dist.~\cite{zhang1989simple} & $\downarrow$   \\ \cline{2-4} 
 & \multirow{3}{*}{Semantic} & MoverScore~\cite{zhao-etal-2019-moverscore} & $\uparrow$   \\
 &  & USE Sim.~\cite{cer-etal-2018-universal} & $\uparrow$  \\
 &  & SBERT Sim.~\cite{reimers2019sentence} & $\uparrow$  \\ \hline
\multirow{6}{*}{Diversity} & \multicolumn{1}{l}{\multirow{3}{*}{Lexical}} & Self-BLEU~\cite{zhu2018texygen} & $\downarrow$   \\
 & \multicolumn{1}{l}{} & Distinct-n~\cite{li-etal-2016-diversity} & $\uparrow$   \\
 & \multicolumn{1}{l}{} & Levenshtein Dist.~\cite{levenshtein1966binary} & $\uparrow$   \\ \cline{2-4} 
 & \multirow{2}{*}{Semantic} & SBERT sim.~\cite{reimers2019sentence} & $\downarrow$   \\
 &  & BERTScore~\cite{zhang2019bertscore} & $\downarrow$  \\
 \hline
\multicolumn{2}{l}{\multirow{2}{*}{Fluency}} & Likelihood Rate~\cite{salazar-etal-2020-masked} & ($\rightarrow 1$)   \\
\multicolumn{2}{l}{} & Perplexity Score~\cite{radford2019language} & $\downarrow$   \\ \hline
\multicolumn{2}{l}{\multirow{2}{*}{Model Performance}} & Accuracy / F1-Score & $\uparrow$   \\
\multicolumn{2}{l}{} &  Std of accuracy / F1-score in multiple runs & $\downarrow$   \\ 
\Xhline{1.2pt}
\end{tabular}}
\vspace{-0.4cm}
\end{table}


\begin{table*}[tb]
\centering
\caption{Summary of CFE generation based on manual annotation.}
\label{tab:human paper list}
\setlength{\extrarowheight}{7pt}
\resizebox{\linewidth}{!}{%
\begin{tabular}{l|l|l|l} 
\hline
\multicolumn{1}{c|}{Method} & Task: Dataset & Annotators &  Project Link \\ 
\hline
\cite{kaushik2019learning} & SA: IMDB; NLI: SNLI & Crowd worker & \url{https://github.com/acmi-lab/counterfactually-augmented-data} \\ 
\hdashline[1pt/1pt]
\cite{qin2019counterfactual} & SR: TIMETRAVEL & Crowd worker & \url{https://github.com/qkaren/Counterfactual-StoryRW} \\ 
\hdashline[1pt/1pt]
\cite{khashabi-etal-2020-bang} & QA: BOOLQ & Master worker & \url{https://github.com/allenai/natural-perturbations} \\ 
\hdashline[1pt/1pt]
\cite{gardner-etal-2020-evaluating} & \begin{tabular}[c]{@{}l@{}}SA: IMDB;\\NLI: PERSPECTRUM; \\QA: DROP,QUOREF,ROPES,\\  \ \  \ \ \   \ \ \ MC-TACO, BOOLQ;\\RE: MATRES\end{tabular} & Domain expert & \url{https://allennlp.org/contrast-sets} \\ 
\hdashline[1pt/1pt]
\cite{sathe-etal-2020-automated} & NLI: WIKIFACTCHECK & Crowd worker & \url{http://github.com/WikiFactCheck-English} \\ 
\hdashline[1pt/1pt]
\cite{samory2021call} & Sexism: CMSB & Crowd worker & \url{https://doi.org/10.7802/2251} \\
\hdashline[1pt/1pt]
\cite{sha-etal-2021-controlling}&QA: WIKIBiOCTE & linguistics & \url{https://sites.google.com/view/control-text-edition/home} \\
\hline
\end{tabular}
}

\vspace{1cm}

\centering
\caption{Summary of CFE generation based on joint learning-based generation.  `MP' means model performance. For the unique formula in validity evaluation, we list the models applied. Symbols \ding{55} and  \ding{51} depict ``not included'' and ``included'' respectively. Papers are organized chronologically.}
\label{tab:gradient paper list}
\setlength{\extrarowheight}{5pt}
\resizebox{\linewidth}{!}{%
\begin{tabular}{l|l|ll|lllll} 
\hline
\multicolumn{1}{c|}{\multirow{2}{*}{Method}}& \multicolumn{1}{c|}{\multirow{2}{*}{Task}} & \multicolumn{2}{c|}{Solution} &   \multicolumn{5}{c}{Evaluation} \\ 
\cline{3-4}\cline{5-9}
 &  & \multicolumn{1}{c}{Objectives}   & \multicolumn{1}{c|}{Filter} & Validity & Diversity & Proximity & Fluency & MP \\ 
\hline
GYC \cite{madaan2021generate} & CLASS &\makecell[l]{Val.+Pro.+ Div.} & \ding{55}  & XL-Net & BERTScore $\downarrow$ & \makecell[l]{Syntax Dist. $\downarrow$\\SBERT Sim. $\uparrow$} & Human & \ding{51} \\ 
\hdashline[1pt/1pt]
CounterfactualGAN \cite{robeer2021generating} & CLASS & Val.+Pro. &  Val. & BERT &  1-USE $\uparrow$  & \ding{55} & Human & \ding{55} \\ 
\hdashline[1pt/1pt]
\citet{hu2021causal} & CLASS & Val.+Pro.+Flu. & \ding{55}  & \makecell[l]{GPT-2\\Human }& Distinct-2 $\uparrow$ & BLEU $\uparrow$ & GPT-2 Perplexity $\downarrow$ & \ding{55} \\ 
\hdashline[1pt/1pt]
GradualCAD \cite{jung2022counterfactual} & CLASS &  Val.+Pro. &  \ding{55}  & \ding{55}  & \ding{55}  & \ding{55}  & \ding{55}  & \ding{51} \\ 
\hdashline[1pt/1pt]
CASPer \cite{madaan2023counterfactual} & CLASS & Val.+Flu.+Pro. & \ding{55}  & \ding{55}  & BLEU $\downarrow$ & SBERT Sim. $\uparrow$ & GPT-2 Perplexity $\downarrow$ & \ding{51} \\ 
\hdashline[1pt/1pt]
MATTE \cite{yan2024counterfactual} & SA & Val.+Pro.+Flu. & \ding{55}  & CNN & Diversity-2 $\uparrow$ & \makecell[l]{BLEU $\uparrow$ \\ Human} & \makecell[l]{GPT-2 Perplexity $\downarrow$ \\ Human} & \ding{51} \\ 
\hline
\end{tabular}
}

\vspace{1cm}

\centering
\caption{Summary of CGE generation based on LLM prompting. `MP' represents model performance, and for the unique formula in validity evaluation, we list the models applied. Symbols \ding{55} and  \ding{51} depict ``not included'' and ``included'' respectively. Papers are listed chronologically.}
\label{tab:llm paper list}
\setlength{\extrarowheight}{7pt}
\resizebox{\linewidth}{!}{%
\begin{tabular}{l|l|ll|lllll} 
\hline
\multicolumn{1}{c|}{\multirow{2}{*}{Method}}& \multicolumn{1}{c|}{\multirow{2}{*}{Task}} & \multicolumn{2}{c|}{Solution} &   \multicolumn{5}{c}{Evaluation} \\ 
\cline{3-4}\cline{5-9}
 &  & \multicolumn{1}{c}{Prompting}   & \multicolumn{1}{l|}{Filter} & Validity & Diversity & Proximity & Fluency & MP \\ 
\hline
CORE \cite{dixit-etal-2022-core} & CLASS & ICL  & \ding{55}  & Human & \makecell[l]{Self-BLEU $\downarrow$ \\ \#Perturb Type $\uparrow$} & Levenshtein $\downarrow$ & \ding{55}  & \ding{51}  \\ 
\hdashline[1pt/1pt]
DISCO \cite{chen2023disco} & CLASS & ICL  & Val.+Flu.  & Human & \makecell[l]{Self-BLEU $\downarrow$ \\ OTDD $\uparrow$} & \ding{55}  & \ding{55}  & \ding{51}  \\ 
\hdashline[1pt/1pt]
\cite{zhou2023explore} &CLASS & ICL& \ding{55}  & \ding{55}  & \ding{55}  & \ding{55}  & \ding{55}  & \ding{51}  \\ 
\hdashline[1pt/1pt]
\cite{sachdeva-etal-2024-catfood} & QA & \makecell[l]{ICL\\+ CoT}  & Val. & \makecell[l]{FLAN-UL2\\+ GPT-J\\+ GPTNeoX\\+ LLaMA} & \makecell[l]{Self-BLEU $\downarrow$\\Levenshtein $\uparrow$\\SBERT Sim. $\downarrow$\\Semantic Equ.$\downarrow$} & \ding{55}  & \ding{55}  & \ding{51}  \\ 
\hdashline[1pt/1pt]
\cite{gat2023faithful} & CLASS & ICL  & Val. & \ding{55}  & \ding{55}  & \ding{55}  & \ding{55}  & \ding{51}  \\ 
\hdashline[1pt/1pt]
\cite{nguyen2024llms} & CLASS & ICL+CoT  & \ding{55} & BERT & \ding{55}  & Levenshtein $\downarrow$ & GPT-2 Perplexity $\downarrow$ & \ding{51}  \\ 
\hdashline[1pt/1pt]
\cite{li2024prompting} & CLASS & CoT  & Val.+Flu. & \ding{55}  & \ding{55}  & \ding{55}  & \ding{55}  & \ding{51}  \\ 
\hdashline[1pt/1pt]
\cite{bhattacharjee2024llmguided} & CLASS & CoT  & \ding{55} & DistilBERT & \ding{55}  & \makecell[l]{Levenshtein $\downarrow$ \\ USE $\uparrow$} & \ding{55}  & \ding{55}  \\ 
\hdashline[1pt/1pt]
\cite{miao-etal-2024-episodic} & RE & ICL+CoT  & \ding{55} & \ding{55} & \ding{55}  & \ding{55} & \ding{55}  & \ding{51}  \\ 
\hline
\end{tabular}
}
\end{table*}

\begin{table*}[tb]
\centering
\caption{Summary of CFE generation within Identify-and-then-Generate framework. ``W.I.'' means word importance techniques, ``W.S.'' is the word statistic techniques, and ``ALL'' is to leverage all words of a text. Papers are listed chronologically. }
\label{tab:ig paper list}
\setlength{\extrarowheight}{3pt}
\resizebox{\linewidth}{!}{%
\begin{tabular}{l|l|lll|lllll} 
\hline
\multicolumn{1}{c|}{\multirow{2}{*}{Method}} & \multicolumn{1}{c|}{\multirow{2}{*}{Task}} & \multicolumn{3}{c|}{Solution}  & \multicolumn{5}{c}{Evaluation} \\ 
\cline{3-5}\cline{6-10}
\multicolumn{1}{c|}{} & \multicolumn{1}{c|}{} & Identify & Generate & Filter & Validity & Diversity & Proximity & Fluency & MP \\ 
\hline
SEDC~\cite{martens2014explaining} & CLASS & W.I. & Delete & \ding{55}  & SVM & \ding{55}  & \#Delete Word $\downarrow$  & \ding{55}  & \ding{55}  \\ 
\hdashline[1pt/1pt]
DeleteAndRetrieve~\cite{li-etal-2018-delete} & CLASS & W.S. & \makecell[l]{Retrieve \\Semantic Edit\\Open Infilling} & Flu. & \makecell[l]{Bi-LSTM \\Human }& \ding{55}  & \makecell[l]{BLEU $\uparrow$\\Human} & Human & \ding{55}  \\ 
\hdashline[1pt/1pt]
AC-MLM~\cite{ijcai2019-732} & SA & W.S.+W.I. & MLM Infilling & \ding{55}  & \makecell[l]{Bi-LSTM \\Human } & \ding{55}  & BLEU $\uparrow$ & Human & \ding{55}  \\ 
\hdashline[1pt/1pt]
PAWS~\cite{zhang-etal-2019-paws} & NLI & Parser & 
MLM Infilling & Val.  & Human & \ding{55}  & \ding{55}  & Human & \ding{51}  \\ 
\hdashline[1pt/1pt]
Tag-and-Generate~\cite{madaan-etal-2020-politeness} & SA & W.S. & MLM Infilling & \ding{55}  & \makecell[l]{AWD-LSTM\\Human} & \ding{55}  & \makecell[l]{BLEU $\uparrow$\\ ROUGE $\uparrow$\\ METEOR $\uparrow$\\Human} & Human & \ding{55}  \\ 
\hdashline[1pt/1pt]
MASKER~\cite{malmi-etal-2020-unsupervised} & CLASS & W.I. & MLM Infilling & \ding{55}  & BERT & \ding{55}  & BLEU $\uparrow$ & \ding{55}  & \ding{55}  \\ 
\hdashline[1pt/1pt]
LIT~\cite{li-etal-2020-linguistically} & NLI & Parser & Syntax Edit & Flu. & Human & \ding{55}  & \ding{55}  & Human & \ding{51}  \\ 
\hdashline[1pt/1pt]
CheckList~\cite{ribeiro-etal-2020-beyond} & CLASS & Parser & \makecell[l]{MLM Infilling\\ Semantic Edit} & \ding{55}  & \ding{55}  & \ding{55}  & \ding{55}  & \ding{55}  & \ding{51}  \\ 
\hdashline[1pt/1pt]
REP-SCD~\cite{yang-etal-2020-generating} & CLASS & W.I. & \textcolor{black}{\makecell[l]{MLM Infilling}} & \ding{55}  & \ding{55}  & \ding{55}  & \ding{55}  & Human & \ding{51}  \\ 
\hdashline[1pt/1pt]
\cite{ramon2020comparison} & CLASS & W.I. & Delete & \ding{55}  & SVM & \ding{55}  & \#Delete Word $\downarrow$  & \ding{55}  & \ding{55}  \\ 
\hdashline[1pt/1pt]
\cite{asai-hajishirzi-2020-logic} &QA	& Parser & Semantic Edit &	Val.	& \ding{55}  &	\ding{55}  & \ding{55} 	& \ding{55} 	& \ding{51} \\
\hdashline[1pt/1pt]
LEWIS~\cite{reid-zhong-2021-lewis} & SA & W.I. & MLM Infilling & Val. & \makecell[l]{RoBERTa\\Human}  & \ding{55}  & \makecell[l]{BLEU $\uparrow$\\ BERTScore $\uparrow$\\ Human} & Human & \ding{51}  \\ 
\hdashline[1pt/1pt]
Polyjuice~\cite{wu-etal-2021-polyjuice} & CLASS & Parser & MLM Infilling & Flu. & Human & Self-BLEU $\downarrow$ & \makecell[l]{Levenshtein $\downarrow$\\ Syntax Dist. $\downarrow$} & Human & \ding{51}  \\ 
\hdashline[1pt/1pt]
MiCE~\cite{ross-etal-2021-explaining} & CLASS & W.I. & MLM Infilling & \ding{55}  & RoBERTa  & \ding{55}  & Levenshtein $\downarrow$ & T5 Likelihood & \ding{55}  \\ 
\hdashline[1pt/1pt]
\cite{wang2021robustness} & SA & W.I. & Semantic Edit & \ding{55}  & \ding{55}  & \ding{55}  & \ding{55}  & \ding{55}  & \ding{51}  \\ 
\hdashline[1pt/1pt]
CrossAug~\cite{lee2021crossaug} & NLI & W.I. & \makecell[l]{Open Infilling \\+Syntactic Edit} & \ding{55}  & \ding{55}  & \ding{55}  & \ding{55}  & \ding{55}  & \ding{51}  \\ 
\hdashline[1pt/1pt]
SentimentCAD~\cite{yang-etal-2021-exploring} & SA & W.I. & \makecell[l]{\textcolor{black}{MLM Infilling}} & Pro. & \ding{55}  & \ding{55}  &  \ding{55}& \ding{55}  & \ding{51}  \\ 
\hdashline[1pt/1pt]
\cite{longpre-etal-2021-entity} & QA & Parser & Syntactic Edit & \ding{55}  & Human  & \ding{55}  & \ding{55}  & Human & \ding{51}  \\ 
\hdashline[1pt/1pt]
SMG~\cite{sha-etal-2021-controlling} & QA & W.I. & MLM Infilling & \ding{55}  & Human & \ding{55}  & BLEU $\uparrow$ & \makecell[l]{ KNM Perplexity $\downarrow$\\ Human} & \ding{55}  \\ 
\hdashline[1pt/1pt]
KACE~\cite{chen2021kace} & NLI & W.I. & Semantic Edit & \makecell[l]{Val.+Pro.\\+Div.} & Human & \ding{55}  & Human  & \ding{55} & \ding{51}  \\ 
\hdashline[1pt/1pt]
RCDA~\cite{chen-etal-2021-reinforced} & SA & Parser & Semantic Edit & \ding{55}  & \ding{55}  & Distinct-2 $\uparrow$ & \ding{55}  &  \ding{55} & \ding{51}  \\ 
\hdashline[1pt/1pt]
PARE~\cite{ross2021learning} & CLASS & Parser & Semantic Edit & \ding{55}  & \ding{55} & \ding{55}  & \ding{55}  & \ding{55}  & \ding{51}  \\ 
\hdashline[1pt/1pt]
CLOSS~\cite{fern-pope-2021-text} & CLASS & ALL & Heuristic Search & \ding{55}  & \makecell[l]{RoBERTa\\ BERT} & \ding{55}  & \makecell[l]{BLEU $\uparrow$\\ Edit Fraction $\downarrow$} & GPT-J Perplexity $\downarrow$ & \ding{55}  \\ 
\hdashline[1pt/1pt]
Sketch-and-Customize~\cite{hao2021sketch} & SR & W.I. & MLM Infilling & \ding{55}  & Human  & \ding{55}  & \makecell[l]{BLEU $\uparrow$\\ ROUGE-L $\uparrow$\\  Human} & \ding{55}  & \ding{55}  \\ 
\hdashline[1pt/1pt]
Tailor~\cite{ross-etal-2022-tailor} & CLASS & Parser & MLM Infilling & Flu. & Human & Edit Fraction $\uparrow$ & F1 Score $\downarrow$ & \makecell[l]{GPT-2 Likelihood \\ Human }& \ding{51}  \\ 
\hdashline[1pt/1pt]
RGF~\cite{paranjape-etal-2022-retrieval} & QA & \ding{55}  & \makecell[l]{Retrieved Context\\+ Open Infilling} & Val. + Pro. & \makecell[l]{T5\\Human} & \#Edit Type $\uparrow$  & Levenshtein $\downarrow$ & Human & \ding{51}  \\ 
\hdashline[1pt/1pt]
BPB~\cite{geva2022break} & QA & Parser & \makecell[l]{Syntactic Edit\\Open Infilling} & \ding{55}  & Human &\ding{55} & \ding{55}  & \ding{55}  & \ding{51}  \\ 
\hdashline[1pt/1pt]
AutoCAD~\cite{wen2022autocad} & CLASS & W.I. & MLM Infilling & Val. & RoBERTa & Distinct-n $\uparrow$ & \ding{55}  & \ding{55}  & \ding{51}  \\ 
\hdashline[1pt/1pt]
CAT~\cite{chemmengath-etal-2022-cat} & CLASS & W.I. & MLM Infilling & \makecell[l]{Val.+Div.\\+Flu.+Pro.} & \makecell[l]{RoBERTa\\Human} & \ding{55}  & \makecell[l]{Levenshtein  $\downarrow $\\BERTScore  $\uparrow $} & GPT-2 Likelihood & \ding{55}  \\ 
\hdashline[1pt/1pt]
NeuroCFs~\cite{howard-etal-2022-neurocounterfactuals} & SA & Parser & Open Infilling & \ding{55} & \ding{55}  & Distinct-n  $\uparrow $ & \makecell[l]{Levenshtein  $\downarrow $\\BLEU-2  $\uparrow $\\MoverScore  $\uparrow $} & GPT-J Perplexity  $\downarrow $ & \ding{51}  \\ 
\hdashline[1pt/1pt]
DoCoGen~\cite{calderon-etal-2022-docogen} & DA & W.S. & MLM Infilling & Val.+Pro. & Human & \ding{55}  & Human & Human & \ding{51}  \\ 
\hdashline[1pt/1pt]
EDUCAT~\cite{chen2022unsupervised} & SR & W.I. & \makecell[l]{MLM Infilling} & \ding{55}  & \makecell[l]{RoBERTa\\Human} & \ding{55}  & \makecell[l]{BLEU  $\uparrow $\\BERTScore  $\uparrow $\\Human} & \ding{55}  & \ding{55}  \\ 
\hdashline[1pt/1pt]
RACE~\cite{zhu-etal-2023-explain} & NLI & W.I. & \makecell[l]{Syntactic Edit\\+Open Infilling} & Val.+Pro. & \makecell[l]{RoBERTa\\Human} & \makecell[l]{ $1/{\textrm{BLEU}}$  $\uparrow $\\Human} & \makecell[l]{MoverScore $\uparrow $\\Human}  & \makecell[l]{GPT-2 Perplexity $\downarrow $ \\ Human}& \ding{51}  \\ 
\hdashline[1pt/1pt]
RELITC~\cite{betti2023relevance} & CLASS & W.I. & MLM Infilling & \ding{55}  & RoBERTa & \ding{55}  & \makecell[l]{Levenshtein $\downarrow $ \\BLEU $\uparrow$\\SBERT Sim $\uparrow$\\Mask Fraction$\downarrow$} & GPT-2 Likelihood & \ding{51}  \\ 
\hdashline[1pt/1pt]
CREST~\cite{treviso-etal-2023-crest} & CLASS & W.I. & MLM Infilling & \ding{55}  & \makecell[l]{RoBERTa\\Human} & Self-BLEU  $\downarrow $ & Levenshtein  $\downarrow $ & \makecell[l]{GPT-2 Perplexity  $\downarrow$\\human} & \ding{51}  \\ 
\hdashline[1pt/1pt]
CoCo~\cite{zhang2023towards} & RE & Parser & Syntax Edit & Val. & \makecell[l]{PA-LSTM\\AGGCN\\R-BERT}  & \ding{55}  & \ding{55}  & \ding{55}  & \ding{51}  \\ 
\hdashline[1pt/1pt]
SCENE~\cite{fu-etal-2023-scene} & QA & Random & MLM Infilling & Val. & \ding{55}  & \ding{55}  & \ding{55}  & \ding{55}  & \ding{51}  \\ 
\hdashline[1pt/1pt]
CCG~\cite{miao2023generating} & RE & W.I.+Parser & MLM Infilling & Flu.+Val.  & Human  & \ding{55}  & Human & \textit{Grammarly} Tool & \ding{51}  \\ 
\hdashline[1pt/1pt]
Remask~\cite{hong-etal-2023-robust} & DA & W.S.+W.I. & MLM Infilling & \ding{55}  & Human & \ding{55}  & Human & Human & \ding{51}  \\ 
\hdashline[1pt/1pt]
CLICK~\cite{li2023click} & SR & W.I. & MLM Infilling & \ding{55}  & RoBERTa  & \ding{55}  & \makecell[l]{BLEU  $\uparrow$\\BERTScore $\uparrow$} & \ding{55}  & \ding{55}  \\ 
\hdashline[1pt/1pt]
TCE-Search~\cite{gilo2024general} & CLASS & \textcolor{black}{\makecell[l]{W.I.}} & \textcolor{black}{Heuristic Search} & Flu. & \makecell[l]{RoBERTa\\Human } & \ding{55}  & \makecell[l]{Levenshtein  $\downarrow $\\Syntax Dist.  $\downarrow $\\SBERT Sim.  $\uparrow $} & \makecell[l]{GPT-2 Likelihood\\Human} & \ding{55}  \\ 
\hdashline[1pt/1pt]
\cite{wu2024novel} & SA & W.I. & MLM Infilling & Val. & \ding{55}  & \ding{55}  & \ding{55}  & \ding{55}  & \ding{51}  \\ 
\hdashline[1pt/1pt]
CEIB~\cite{chang2024counterfactual} & SA & Random & MLM Infilling & Val. & \ding{55} & \ding{55}  & \ding{55}  & \ding{55}  & \ding{51}  \\
\hdashline[1pt/1pt]
\cite{ijcai2024p721} & SR & W.I. & Open Infilling & \ding{55} & \makecell[l]{FactScore$\uparrow $\\Human} & \ding{55} & \makecell[l]{human \\ ROUGH $\uparrow$\\BERTScore $\uparrow$\\BERT-FT $\uparrow$\\WMS $\uparrow$} & \makecell[l]{NSPScore$\uparrow $\\Human}  & \ding{51}  \\
\hline
\end{tabular}
}
\end{table*}

\clearpage

\begin{figure*}[htbp]
\centering
    \includegraphics[scale=0.22]{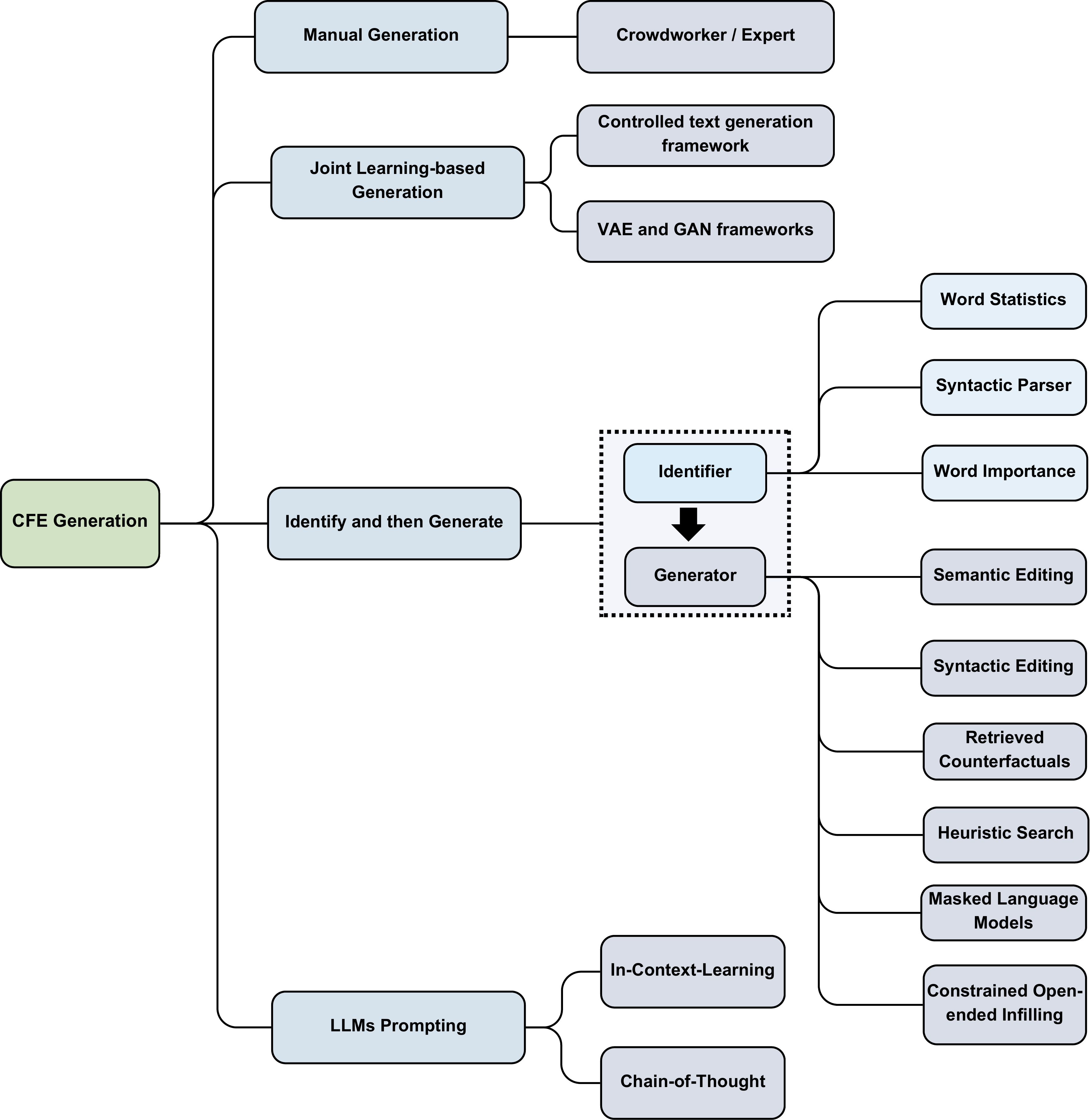}
    \caption{The complete taxonomy proposed for existing literature on natural language counterfactual generation.}
    \label{fig:full_tree}
\end{figure*}

\end{document}